\documentclass[pdflatex,sn-mathphys-num]{sn-jnl}

\usepackage{xcolor}
\usepackage{float}
\usepackage{listings}
\usepackage{multirow}   
\usepackage{comment}
\usepackage{booktabs}
\usepackage{amssymb} 

\definecolor{codegreen}{rgb}{0,0.6,0}
\definecolor{codegray}{rgb}{0.5,0.5,0.5}
\definecolor{codepurple}{rgb}{0.58,0,0.82}
\definecolor{backcolour}{rgb}{0.95,0.95,0.92}

\lstdefinestyle{mystyle}{
    backgroundcolor=\color{backcolour},   
    commentstyle=\color{codegreen},
    keywordstyle=\color{magenta},
    numberstyle=\tiny\color{codegray},
    stringstyle=\color{codepurple},
    basicstyle=\footnotesize\ttfamily,
    breakatwhitespace=false,         
    breaklines=true,                 
    captionpos=b,                    
    keepspaces=true,                 
    numbers=left,                    
    numbersep=5pt,                  
    showspaces=false,                
    showstringspaces=false,
    showtabs=false,                  
    tabsize=2,
    frame=single,
    rulecolor=\color{black!30},
}
\usepackage{array}
\usepackage{ragged2e}
\usepackage{seqsplit} 
\usepackage{algorithm}
\usepackage{algorithmicx}
\usepackage{algpseudocode}
\usepackage{longtable} 
\usepackage{booktabs} 
\usepackage{array}    
\usepackage{tabularx}
\usepackage[table]{xcolor}
\usepackage{graphicx} 
\usepackage{amsmath}
\usepackage{colortbl}
\usepackage{colortbl}    
\definecolor{lightgray}{gray}{0.9} 
\definecolor{gray}{gray}{.8}
\usepackage{xcolor}      
\usepackage{listings}    

\usepackage{graphicx}%
\usepackage{tikz}
\usetikzlibrary{shapes.geometric, arrows.meta, positioning, fit, backgrounds, shadows.blur}
\usepackage{multirow}%
\usepackage{amsmath,amssymb,amsfonts}%
\usepackage{amsthm}%
\usepackage{mathrsfs}%
\usepackage[title]{appendix}%
\usepackage{xcolor}%
\usepackage{textcomp}%
\usepackage{manyfoot}%
\usepackage{booktabs}%
\usepackage{algorithm}%
\usepackage{algorithmicx}%
\usepackage{algpseudocode}%
\usepackage{listings}%

\theoremstyle{thmstyleone}%
\theoremstyle{thmstyletwo}%

\theoremstyle{thmstylethree}%

\begin{document}

\title[Article Title]{A Multi-level Analysis of Factors Associated with Student Performance: A Machine Learning Approach to the SAEB Microdata}

\author*[1]{\fnm{Rodrigo} \sur{Tertulino}}\email{rodrigo.tertulino@ifrn.edu.br}

\author[1]{\fnm{Laércio} \sur{Alencar}}\email{lapea@ifrn.edu.br}

\affil*[1]{\orgdiv{IFRN}, \orgname{Federal Institute of Education, Science, and Technology of Rio Grande do Norte}, \orgaddress{\street{Raimundo Firmino de Oliveira}, \city{Mossoró}, \postcode{59628-330}, \state{RN}, \country{Brazil}}}

\abstract{Identifying the factors that influence student performance in basic education is a central challenge for formulating effective public policies in Brazil. The present work introduces a multi-level machine learning approach to classify the proficiency of 9th-grade and high school students using microdata from the System of Assessment of Basic Education (SAEB). Our model uniquely integrates four data sources: student socioeconomic characteristics, teacher professional profiles, school indicators, and principal management profiles. A comparative analysis of four ensemble algorithms confirmed the superiority of a Random Forest model, which achieved 90.2\% accuracy and an Area Under the Curve (AUC) of 96.7\%. To move beyond prediction, we applied Explainable AI (XAI) using SHAP, which revealed that the school's average socioeconomic level is the most dominant predictor, demonstrating that systemic factors have a greater impact than individual characteristics in isolation. The primary conclusion is that academic performance is a systemic phenomenon deeply tied to the school's ecosystem. The proposed framework provides a data-driven, interpretable tool to inform policies that promote educational equity by addressing disparities across schools.}

\keywords{Student Performance Prediction, Machine Learning, Educational Data Mining, SAEB, Multi-level Analysis, Socioeconomic Factors, Brazilian Basic Education.}



\maketitle

\section{Introduction}

Identifying the determinants of academic success in basic education represents a central challenge for educational research and policymaking, particularly in a country as vast and socioeconomically heterogeneous as Brazil \citep{ISSAH2023100204}. A systemic approach is crucial, as student performance is influenced by a complex interplay of factors spanning multiple analytical levels. Following the framework proposed by \citet{barragan2024complexities}, we classify these influences into four interrelated categories: (1) \textbf{individual factors}, encompassing student-level characteristics such as study habits and demographic background; (2) \textbf{academic factors}, related to the instructional process, such as teacher qualifications and pedagogical practices; (3) \textbf{socioeconomic factors}, capturing the family and community context of students; and (4) \textbf{institutional factors}, concerning school management, leadership, and organizational characteristics. Throughout this study, we use the term \textbf{systemic factors} to collectively refer to variables at the school, teacher, and principal levels, that is, categories (2), (3) at the school level, and (4), as opposed to purely individual student-level characteristics. The System of Assessment of Basic Education (SAEB), conducted by the National Institute for Educational Studies and Research Anísio Teixeira (INEP)~\citep{INEP2025SAEB}, provides a rich, multi-level dataset uniquely suited for such an analysis. The public availability of its anonymized microdata enables the research community to investigate the intricate relationships between student proficiency and a wide array of contextual factors, from socioeconomic backgrounds to school infrastructure and teacher profiles. Consequently, the SAEB microdata is an essential resource for data-driven research aimed at informing and evaluating educational policies in the country \citep{10.5555/3295222.3295230, Mazoni_Oliveira_2023}.

While traditional statistical methods are common, the Educational Data Mining (EDM) paradigm offers powerful tools for uncovering complex, non-linear patterns from such data \citep{romero2010educational}. Furthermore, we demonstrate that by interpreting the model's classification results with XAI techniques, our method provides data-driven insights for educators and policymakers~\citep{10.1145/3660853.3660877}. The primary objective of this research is thus to develop and evaluate a multi-level machine learning model to identify the key systemic factors associated with the academic performance of 9th-grade and high school students, using the SAEB microdata. Building on this perspective, the study shifts its analytical focus from purely individual student interventions to the systemic determinants that shape educational outcomes in Brazilian basic education. To guide this investigation, the following research questions (RQs) were formulated:

\begin{itemize}[label=\textbf{RQ}]
    \item \textbf{RQ1:} How effectively can machine learning models analyze the vast, multi-level SAEB microdata to identify the most significant predictors of student academic performance?
    \item \textbf{RQ2:} What is the relative importance of individual and academic characteristics compared to systemic, socioeconomic, and institutional factors in explaining student achievement?
    \item \textbf{RQ3:} What data-driven insights can be inferred from the final predictive model to help educators and policymakers address the primary factors influencing educational outcomes in Brazil?
\end{itemize}

To address these research questions, this study employs and compares four powerful, tree-based ensemble models, Random Forest \citep{breiman2001random}, XGBoost \citep{chen2016xgboost}, LightGBM \citep{ke2017lightgbm}, and CatBoost \citep{dorogush2018catboost}, selected for their high performance on tabular data. Moreover, the article makes the following contributions to the field of Educational Data Mining and public policy analysis:

\begin{enumerate}[label=\textbf{C\arabic*.}]
    \item We design and implement a novel multi-level predictive model by systematically integrating four distinct data sources from the national SAEB assessment: student socioeconomic profiles, teacher professional data, consolidated school indicators, and principal management profiles.
    \item We conduct a rigorous comparative evaluation of powerful tree-based ensemble models, demonstrating that our final Random Forest model achieves 90.2\% predictive accuracy and an Area Under the Curve (AUC) of 96.7\%.
    \item We provide a data-driven feature importance analysis that unequivocally identifies the school's average socioeconomic level as the most dominant predictor, demonstrating that systemic, school-level factors have a greater impact on performance than isolated individual characteristics.
    \item We offer actionable, evidence-based insights for educational policy, quantifying the systemic nature of academic achievement and highlighting the potential of policies to reduce socioeconomic disparities between schools, thereby fostering greater educational equity.
\end{enumerate}

In addition, this study makes a novel contribution by addressing these questions through the systematic integration and methodologically sound analysis of four distinct data sources. The resulting model provides a realistic measure of predictive accuracy and a nuanced, data-driven perspective on the systemic nature of student achievement, offering actionable insights for educational stakeholders in Brazil.

The remainder of this study is organized as follows: Section~\ref{sec:related_work} reviews related work on educational data mining and performance prediction. Section~\ref{sec:background} delineates the theoretical foundations guiding the study. Section~\ref{sec:methodology} details our methodology, from data preparation to the machine learning models used. Subsequently, Section~\ref{sec:results} presents the empirical results of our comparative analysis, which are then interpreted through the application of SHAP. Section~\ref{sec:Discussion} discusses the broader implications of our findings in both theoretical and practical contexts. Finally, Section~\ref{sec:conclusion} concludes the paper, summarizing its contributions and outlining directions for future research.

\section{Related Work}
\label{sec:related_work}

Analyzing large-scale educational assessment data has become a cornerstone of educational research. Educational Data Mining (EDM) focuses on developing and applying computational methods to explore data from educational settings, with the overarching goal of understanding and improving student learning processes \citep{papadogiannis2024educational}. A primary focus within EDM has been predicting student outcomes, such as academic performance and dropout risk, to enable timely and targeted interventions \citep{11012110, 10616421}. 

In the Brazilian context, numerous studies have utilized machine learning techniques to address these challenges, primarily in the higher education sector. A significant body of work has applied classification models to predict student dropout using institutional academic records, demonstrating the viability of EDM for identifying at-risk students in the country \citep{sales2015exploiting, carneiro_edm_2022, colpo_slr_2024, melo_semisupervised_2023}. These studies establish a strong precedent for using computational models to tackle critical issues in Brazilian education. Globally, predicting student performance is a well-established area of research. Early identification of students likely to underperform enables educators to provide targeted support, laying the groundwork for adaptive and personalized learning platforms \citep{martinez2024early, 10434888, 10734474}. 

A wide range of machine learning algorithms has been employed for the purpose, with increasing attention given to comparisons between traditional models and more sophisticated approaches, such as deep learning \citep{pereira_deeplearning_2020}. More recently, the field has shifted not only towards accurate prediction but also towards interpretability. The emergence of Explainable Artificial Intelligence (XAI) in education aims to elucidate the decision-making process of predictive models, providing clear insights into which factors are most influential in determining a student's predicted outcome \citep{jang2022practical, ujkani2024course, Mastour2025}. While these studies provide a strong foundation, a research gap remains concerning the application of a multi-level, systemic approach to Brazil's most comprehensive and recent national dataset for basic education.  

Table~\ref{Summary of Related Work} summarizes the reviewed studies, outlining their main contributions and limitations while highlighting the research gap addressed in this work. It provides a concise overview of prior literature and situates the present study within this broader context. By applying a systemic, multi-level machine learning approach to a large-scale national dataset, we extend previous efforts beyond single-institution or single-level analyses. Using robust ensemble models, Random Forest, XGBoost, LightGBM, and CatBoost, we identify the most influential predictors of student performance in Brazilian basic education.

\begingroup
\small
\begin{longtable}{p{0.2\textwidth} p{0.3\textwidth} p{0.4\textwidth}}
\caption{Summary of Related Work}\\
\toprule
\label{Summary of Related Work}
\textbf{Reference} & \textbf{Core Contribution} & \textbf{Limitation / Gap Relevant to Our Work} \\
\midrule
\endfirsthead
\multicolumn{3}{c}%
{{\bfseries \tablename\ \thetable{} (Continued)}} \\
\toprule
\textbf{Reference} & \textbf{Core Contribution} & \textbf{Limitation / Gap Relevant to Our Work} \\
\midrule
\endhead
\bottomrule
\endfoot
\cite{papadogiannis2024educational} & Provides a foundational overview of EDM concepts, methodologies, and applications. & Does not focus on a specific national context or multi-level modeling. \\
\cite{11012110} & Presents a bibliometric analysis of ML in distance education, identifying trends and key research areas. & Focus is on distance education, whereas our study addresses the broader basic education system. \\
\cite{10616421} & Explores the future integration of machine learning in higher education, discussing potential impacts. & Conceptual focus on higher education; lacks an empirical application on a large-scale basic education dataset. \\
\addlinespace
\cite{sales2015exploiting} & A foundational case study on predicting student dropout in a Brazilian public university using academic records. & Focused on dropout, not performance, and limited to a single higher education institution. \\
\cite{carneiro_edm_2022} & Applies EDM to identify and prevent dropout in introductory programming courses in Brazil. & Domain-specific (Computer Science) and focused on a single course level. \\
\cite{colpo_slr_2024} & A systematic review of trends, opportunities, and challenges in EDM for dropout prediction. & Provides context, but is a review, not an empirical study. Highlights the need for more comprehensive models. \\
\cite{melo_semisupervised_2023} & Explores semi-supervised learning to improve dropout prediction with limited labeled data in a Brazilian context. & Focuses on a specific ML technique (semi-supervised) for dropout, not multi-level performance prediction. \\
\addlinespace
\cite{martinez2024early} & Uses engagement, demographic, and performance data for early detection of at-risk students in a US university. & Single-institution study; lacks the national, multi-level scope of SAEB. \\
\addlinespace
\cite{10434888} & Proposes an ML method to support personalized education in flexible learning environments. & Focuses on the application (personalization) rather than the foundational prediction task across a national system. \\
\cite{10734474} & Investigates an adaptive education platform to improve outcomes in higher education. & Platform-specific study; does not perform a broad analysis of factors from a national assessment. \\
\cite{pereira_deeplearning_2020} & Compares deep learning against traditional ML models for early performance prediction in a Brazilian university course. & Limited to a single course and institution; does not integrate systemic, multi-level factors. \\
\addlinespace
\cite{jang2022practical} & A key study on early performance prediction that incorporates XAI to explain model decisions. & Focuses on a specific online learning context, rather than a national, multi-level dataset like SAEB. \\
\cite{ujkani2024course} & Predicts course success and identifies at-risk students using XAI techniques. & Limited to a single university course context. \\
\cite{Mastour2025} & Develops an XAI framework to predict medical students' performance in high-stakes assessments. & Domain-specific (medical education) and institution-focused. \\
\midrule
\textbf{Our Study} & \textbf{Develops a multi-level prediction model by integrating student, teacher, school, and principal data from the national SAEB assessment to identify systemic factors associated with student performance.} & \textbf{Addresses the gap by applying a systemic, multi-level machine learning approach to a recent, large-scale national dataset for Brazilian basic education.} \\
\bottomrule
\end{longtable}
\endgroup

\section{Background: The System of Assessment of Basic Education (SAEB)} 
\label{sec:background}

The System of Assessment of Basic Education (SAEB) is a comprehensive, large-scale evaluation conducted periodically by the National Institute for Educational Studies and Research Anísio Teixeira (INEP), an agency of the Brazilian Ministry of Education. Its primary objective is to assess the quality of basic education in Brazil and identify factors that may impact student performance \citep{franco_2001}. SAEB comprises standardized tests in core subjects, such as Portuguese and Mathematics, and a series of contextual questionnaires administered to students, teachers, school principals, and education secretaries. The resulting microdata provide a rich, multidimensional snapshot of the national educational landscape, making them an invaluable resource for research and policy formulation. For this study, we utilized the microdata from the SAEB administration, focusing on 9th-grade and high school students. The multi-level nature of our analysis was made possible by integrating four distinct, yet interconnected, data sources provided by INEP \citep{INEP2025SAEB}: 

\begin{itemize} 
\item \textbf{STUDENT:} The dataset contains student-level information. It includes proficiency scores in Mathematics and Portuguese, which serve as our target variables, along with responses to a socioeconomic questionnaire. These responses provide critical data on the student's family background, home resources, and personal study habits.    
 \item \textbf{TEACHER:} The dataset captures teacher-level data. It consists of responses to a questionnaire covering the teacher's academic background, professional experience, pedagogical practices, and perceptions of the school climate. Each teacher's record is linked to a specific school and class.     
\item \textbf{SCHOOL:} The source provides school-level aggregated indicators. It contains a calculated index of the school's socioeconomic level, the percentage of teachers with adequate formation, and student participation rates. These variables provide a consolidated view of the school's context and resources.     
\item \textbf{PRINCIPAL:} The dataset contains information from the school principal's questionnaire. It provides insights into the school's management and leadership, including the principal's professional experience, training in school administration, and practices for community and parental engagement. 
\end{itemize} 

By joining these four datasets using common identifiers (e.g., school ID, class ID), we constructed a comprehensive, hierarchical dataset that allows our model to analyze how factors at the student, classroom, school, and management levels are collectively associated with academic performance.

\subsection{Machine Learning Models for Educational Data}

To analyze the complex relationships within the SAEB dataset, this study employs four powerful tree-based ensemble models: Random Forest, XGBoost, LightGBM, and CatBoost. These models were chosen for their high predictive accuracy and ability to handle heterogeneous data types~\citep{technologies13030088}.

\begin{itemize}
    \item \textbf{Random Forest:} The Random Forest algorithm constructs many decision trees during training. For a classification task, each tree in the forest casts a vote for a class, and the class with the most votes becomes the model's final prediction. This ensemble approach corrects for the tendency of individual decision trees to overfit to their training data, resulting in a more robust and generalizable model \citep{breiman2001random}.

    \item \textbf{XGBoost (Extreme Gradient Boosting):} XGBoost is an advanced implementation of the gradient boosting algorithm. Unlike Random Forest, which builds trees independently, XGBoost builds them sequentially. Each new tree is trained to correct the errors made by the previous ones. The sequential, error-correcting process often leads to highly accurate models and has made XGBoost a dominant algorithm in many data science competitions \citep{chen2016xgboost, Arif2023ExploringTP}.

    \item \textbf{LightGBM (Light Gradient Boosting Machine):} LightGBM is another high-performance gradient boosting framework known for its fast training speed and lower memory usage. This efficiency is achieved through techniques such as Gradient-based One-Side Sampling (GOSS) and Exclusive Feature Bundling (EFB), making it highly suitable for large datasets \citep{ke2017lightgbm}.

    \item \textbf{CatBoost (Categorical Boosting):} CatBoost is a gradient boosting algorithm that excels at handling categorical data. It incorporates a novel algorithm for processing categorical features directly, which can reduce the need for extensive preprocessing, such as one-hot encoding, and often improves model accuracy when a dataset contains many categorical variables \citep{dorogush2018catboost}.

\end{itemize}

By comparing the performance of Random Forest against these three distinct gradient boosting implementations, this study comprehensively evaluates their suitability for modeling the systemic factors that influence student performance in the SAEB context.

\subsection{Explainable AI (XAI)}

The primary goal of Explainable AI (XAI) is to make the outputs of machine learning models more interpretable to humans. Interpretability refers to the degree to which a person can understand why a model made a specific prediction~\citep{electronics8080832}. Moreover, this capability is increasingly beneficial as models are used for decision-making in critical areas of people's lives~\citep{MILLER20191}. Various criteria can classify XAI methods. One key distinction is the scope of the explanation, which can be either 
global, seeking to understand the model's behavior as a whole, or local, focusing on explaining a single prediction for an individual instance~\citep{8466590}. Another important criterion is whether a method is model-agnostic, meaning it applies to any machine learning algorithm, or model-specific. Besides that, we selected the SHAP (SHapley Additive exPlanations) framework~\citep{Lundberg}, which provides a theoretically grounded, unified measure of feature importance rooted in cooperative game theory. Although the SHAP framework encompasses both model-agnostic implementations (e.g., KernelSHAP) and model-specific ones, we employed the \texttt{TreeExplainer} variant, which is optimized for tree-based ensemble models. This implementation computes exact Shapley values by leveraging the internal tree structure, offering both computational efficiency and mathematical exactness, two properties essential given the scale of this dataset.

\section{Methodology}
\label{sec:methodology}

The research methodology employed a structured data mining process encompassing data acquisition, multi-level integration, preparation, modeling, and evaluation. The entire workflow was implemented in Python, utilizing libraries such as Pandas for data manipulation and Scikit-learn for machine learning modeling~\citep{10.1145/3689737}. The full pipeline is detailed in Algorithm \ref{alg:full_pipeline}.

\subsection{Data Source and Acquisition}

The primary data source was the official microdata from the System of Assessment of Basic Education (SAEB), publicly available through the National Institute for Educational Studies and Research Anísio Teixeira (INEP) \citep{INEP2025SAEB}. Specifically, this study uses the \textbf{SAEB 2023 edition}, the most recent cycle available at the time of data collection (accessed in January 2025), which represents the first post-pandemic national assessment conducted under a fully restored administration protocol. This edition was selected because it reflects the current state of Brazilian basic education and incorporates the most up-to-date socioeconomic and institutional questionnaires. To construct a comprehensive multi-level model, this study integrated data from both 9th-grade and high school students, using four distinct datasets: \texttt{STUDENT}, \texttt{TEACHER}, \texttt{SCHOOL}, and \texttt{PRINCIPAL}.

Table \ref{tab:dataset_statistics} provides a descriptive overview of the final population analyzed after the data cleaning and integration process. The resulting dataset encompasses a substantial portion of the Brazilian basic education system, comprising \textbf{6,482,168} valid student records. This sample is composed of 4,636,226 (71.52\%) students from the 9th grade and 1,845,942 (28.48\%) students from high school. This large-scale, nationwide scope ensures that the findings are representative and provide a robust foundation for applying machine learning models. For methodological clarity, a complete glossary of all variables used is provided in Appendix~\ref{appendix:Variable Glossary}.

\begin{table}[!ht]
\centering
\caption{Descriptive Statistics of the Final Analyzed Dataset.}
\label{tab:dataset_statistics}
\begin{tabular}{lr}
\toprule
\textbf{Metric} & \textbf{Count} \\
\midrule
\multicolumn{2}{c}{\textbf{Student Distribution}} \\
\midrule
Total Valid Student Records & 6,482,168 \\
- 9th Grade Students & 4,636,226 \\
- High School Students & 1,845,942 \\
\midrule
\multicolumn{2}{c}{\textbf{Contextual Data Overview}} \\
\midrule
Total Unique Schools & 70,151 \\
Total Unique Principals & 73,595 \\
Total Unique Teachers & 290,972 \\
\midrule
\multicolumn{2}{c}{\textbf{Teacher Qualification Distribution}$^{a}$} \\
\midrule
High School & 3,378 \\
Graduation & 70,925 \\
Specialization & 190,517 \\
Masters & 28,094 \\
Doctorate & 4,111 \\
\bottomrule
\end{tabular}
\vspace{2pt}
\begin{minipage}{\linewidth}
\footnotesize $^{a}$The qualification counts sum to 297,025, slightly exceeding the total of 290,972 unique teachers. This occurs because a small fraction of teachers reported more than one completed qualification level in the SAEB questionnaire (e.g., teachers holding both a Specialization and a Master's degree), and each completed level is counted independently in this distribution.
\end{minipage}
\end{table}

\begin{algorithm}
\caption{Full Methodological Pipeline}
\label{alg:full_pipeline}
\begin{algorithmic}[1]
\State \textbf{Input:} Paths to raw data files (\texttt{TS\_ALUNO}, \texttt{TS\_PROFESSOR}, \texttt{TS\_ESCOLA}, \texttt{TS\_DIRETOR})
\State \textbf{Output:} Trained classification model and performance metrics

\Function{FullPipeline}{file\_paths}

    \vspace{2pt}
    \Statex \textbf{--- Phase 1: Data Acquisition and Multi-level Integration ---}
    \State $student, teacher, school, principal \gets \textsc{LoadCSVs}(file\_paths)$
    \State $merged \gets \textsc{Join}(student, teacher, \textbf{on}=[\texttt{ID\_ESCOLA}, \texttt{ID\_TURMA}])$
    \State $merged \gets \textsc{Join}(merged, school, \textbf{on}=\texttt{ID\_ESCOLA})$
    \State $merged \gets \textsc{Join}(merged, principal, \textbf{on}=\texttt{ID\_ESCOLA})$

    \vspace{2pt}
    \Statex \textbf{--- Phase 2: Data Cleaning and Preprocessing ---}
    \State Convert percentage columns from string to float format
    \State Drop identifier columns (\texttt{ID\_ESCOLA}, \texttt{ID\_TURMA}) $\triangleright$ \textit{Prevents data leakage}
    \State Remove rows with true missing values (listwise deletion)
    \State $X \gets \textsc{OneHotEncoding}(merged,\ categorical\_features)$
    \Statex \hspace{2em} $\triangleright$ \textit{Retains ``unknown'' questionnaire categories as valid classes}

    \vspace{2pt}
    \Statex \textbf{--- Phase 3: Target Variable Engineering ---}
    \State $composite\_score \gets \text{mean}(\texttt{PROF\_LP\_SAEB},\ \texttt{PROF\_MT\_SAEB})$
    \State $y \gets [composite\_score > \text{mean}(composite\_score)]$ $\triangleright$ \textit{Binary: 1=Above Avg, 0=Below Avg}

    \vspace{2pt}
    \Statex \textbf{--- Phase 4: Feature Selection ---}
    \State $X_{selected} \gets \textsc{BorutaFeatureSelection}(X, y)$ $\triangleright$ \textit{Retains 44 of 98 features}
    \State $X_{final} \gets \textsc{ExpertCuration}(X_{selected})$ $\triangleright$ \textit{Final 17 features selected}

    \vspace{2pt}
   \State $X_{train}, X_{test}, y_{train}, y_{test} \gets \textsc{TrainTestSplit}(X_{final},\ y,$ \\
       $\qquad test\_size=0.2,\ stratify=y,\ random\_state=42)$
    \Statex \hspace{2em} $\triangleright$ \textit{Stratified split preserves class ratio; seed fixed for reproducibility}
    \State $scaler \gets \textsc{StandardScaler.fit}(X_{train})$ $\triangleright$ \textit{Fitted on train only; prevents leakage}
    \State $X_{train} \gets scaler.transform(X_{train})$;\ $X_{test} \gets scaler.transform(X_{test})$

    \vspace{2pt}
    \Statex \textbf{--- Phase 6: Model Training and Evaluation ---}
    \State $model \gets \textsc{InstantiateClassifier}()$ \Comment{e.g., RandomForestClassifier}
    \State $model.fit(X_{train}, y_{train})$
    \State $metrics_{train} \gets \textsc{Evaluate}(model, X_{train}, y_{train})$
    \State $metrics_{test} \gets \textsc{Evaluate}(model, X_{test}, y_{test})$
    \State \Return $model,\ metrics_{train},\ metrics_{test}$

\EndFunction
\end{algorithmic}
\end{algorithm}

\subsection{Data Preparation and Integration}

Only relevant variables were pre-selected and loaded to optimize computational efficiency. The core of the preparation phase involved integrating the four datasets into a single dataframe using school and class identifiers (\texttt{ID\_ESCOLA} and \texttt{ID\_TURMA}) as join keys. To prevent data leakage, all identifier columns were explicitly removed from the feature set prior to training. Subsequently, records containing \textbf{truly missing values}, defined as blank fields or system-generated null entries in the raw CSV files, were removed using a listwise deletion approach to ensure the models were trained exclusively on complete data instances.

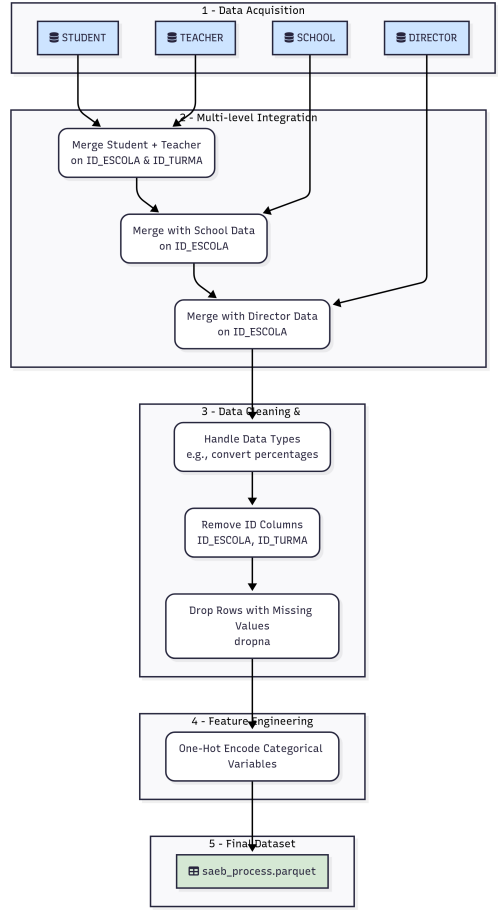
\begin{figure}[h!]
\centering
\begin{tikzpicture}[
    font=\small,
    source/.style={
        rectangle, rounded corners=4pt,
        minimum width=2.6cm, minimum height=1.0cm,
        text centered, text width=2.5cm,
        draw=black!55, thick,
        font=\footnotesize
    },
    phase/.style={
        rectangle, rounded corners=4pt,
        minimum width=11.4cm, minimum height=1.1cm,
        text centered, text width=11.0cm,
        draw=black!60, thick,
        font=\footnotesize
    },
    output/.style={
        rectangle, rounded corners=4pt,
        minimum width=11.4cm, minimum height=0.90cm,
        text centered, text width=11.0cm,
        draw=black!80, very thick,
        font=\footnotesize
    },
    sidelabel/.style={
        font=\scriptsize\bfseries, text=black!55,
        text width=1.5cm, align=center
    },
    arrow/.style={-{Stealth[length=5pt,width=4pt]}, thick, black!60},
]

\node[source, fill=blue!15]   (S1) at (-4.2, 0) {\textbf{STUDENT}\\[1pt]\texttt{TS\_ALUNO}};
\node[source, fill=teal!15]   (S2) at (-1.4, 0) {\textbf{TEACHER}\\[1pt]\texttt{TS\_PROFESSOR}};
\node[source, fill=orange!18] (S3) at ( 1.4, 0) {\textbf{SCHOOL}\\[1pt]\texttt{TS\_ESCOLA}};
\node[source, fill=violet!13] (S4) at ( 4.2, 0) {\textbf{PRINCIPAL}\\[1pt]\texttt{TS\_DIRETOR}};

\node[phase, fill=blue!7] (P1) at (0,-1.55) {%
    \textbf{Phase 1: Multi-level Integration}\\[2pt]
    Join on \texttt{ID\_ESCOLA} \& \texttt{ID\_TURMA} \quad\textbullet\quad 6,482,168 student records};

\draw[arrow] (S1.south) -- ++(0,-0.28) -| ([xshift=-3.8cm]P1.north);
\draw[arrow] (S2.south) -- ++(0,-0.28) -| ([xshift=-1.27cm]P1.north);
\draw[arrow] (S3.south) -- ++(0,-0.28) -| ([xshift= 1.27cm]P1.north);
\draw[arrow] (S4.south) -- ++(0,-0.28) -| ([xshift= 3.8cm]P1.north);

\node[phase, fill=yellow!10] (P2) at (0,-2.9) {%
    \textbf{Phase 2: Data Cleaning}\\[2pt]
    Drop \texttt{ID\_ESCOLA}, \texttt{ID\_TURMA} (prevent leakage) \quad\textbullet\quad Listwise deletion of missing values};

\draw[arrow] (P1.south) -- (P2.north);

\node[phase, fill=orange!7] (P3) at (0,-4.25) {%
    \textbf{Phase 3: Feature Engineering}\\[2pt]
    One-Hot Encoding (incl.\ \texttt{\_F} unknown categories) \quad\textbullet\quad StandardScaler normalization};

\draw[arrow] (P2.south) -- (P3.north);

\node[phase, fill=red!6] (P4) at (0,-5.6) {%
    \textbf{Phase 4: Target Variable Engineering}\\[2pt]
    Binary label: \textit{Above Average} vs.\ \textit{Below Average} \quad\textbullet\quad Threshold = dataset mean};

\draw[arrow] (P3.south) -- (P4.north);

\node[phase, fill=purple!6] (P5) at (0,-6.95) {%
    \textbf{Phase 5: Feature Selection (Boruta)}\\[2pt]
    98 candidates \;\textrightarrow\; 44 confirmed \;\textrightarrow\; \textbf{17 final features} (expert curation)};

\draw[arrow] (P4.south) -- (P5.north);

\node[phase, fill=cyan!7] (P6) at (0,-8.3) {%
    \textbf{Phase 6: Stratified Train / Test Split}\\[2pt]
    \textbf{80\% Training set} (5,185,734 records)
    \quad\textbullet\quad
    \textbf{20\% Test set} (1,296,434 records)};

\draw[arrow] (P5.south) -- (P6.north);

\node[output, fill=green!12] (OUT) at (0,-9.55) {%
    \textbf{Final Processed Dataset: Ready for Modelling}\\[2pt]
    Random Forest \quad XGBoost \quad LightGBM \quad CatBoost};

\draw[arrow] (P6.south) -- (OUT.north);

\node[sidelabel] at (6.6,  0.00) {Data\\Sources};
\node[sidelabel] at (6.6, -1.55) {Integration};
\node[sidelabel] at (6.6, -2.90) {Cleaning};
\node[sidelabel] at (6.6, -4.25) {Encoding\\Scaling};
\node[sidelabel] at (6.6, -5.60) {Labelling};
\node[sidelabel] at (6.6, -6.95) {Feature\\Selection};
\node[sidelabel] at (6.6, -8.30) {Splitting};
\node[sidelabel] at (6.6, -9.55) {Output};

\end{tikzpicture}
\caption{The multi-level data preprocessing pipeline, from raw data acquisition to the final processed dataset ready for model training.}
\label{fig:preprocessing_flowchart}
\end{figure}

\textbf{Important distinction regarding questionnaire response categories:} It is necessary to clarify that the SAEB questionnaire includes a response option labeled ``unknown'' or ``not informed'' (coded as category \texttt{F} in variables such as parental education, \texttt{TX\_RESP\_Q08}, and \texttt{TX\_RESP\_Q09}). This response category is a \textit{valid, substantive answer} recorded in the INEP dataset, not a missing value in the technical sense. Students who selected ``I don't know'' or ``not applicable'' for their parents' education level constitute a distinct, potentially disadvantaged subgroup. Consequently, after listwise deletion of true missing values, these ``unknown'' category entries (\texttt{\_F}) were retained and encoded as a separate class through One-Hot Encoding. The SHAP analysis reported in Section~\ref{sec:results} correctly identifies the negative predictive effect of this category, reflecting the socioeconomic condition of students who lack information about their parents' educational backgrounds, which is itself a meaningful signal of disadvantage~\citep{Delena2025}. The multi-step workflow is visually summarized in Figure \ref{fig:preprocessing_flowchart}.

\subsection{The School Socioeconomic Level Indicator}
\label{sec:nse_indicator}

A central variable in this study is the School Socioeconomic Level Indicator (\texttt{NIVEL\_SOCIO\_ECONOMICO}), which requires precise characterization to avoid interpretive ambiguity. According to the official technical note published by INEP \citep{inep2021nota}, this indicator is constructed \textbf{exclusively from student questionnaire responses}: specifically, it aggregates information on parental education levels and the presence of household assets (e.g., number of books at home, internet access, computers, and other goods). A continuous socioeconomic score is first generated for each student based on these responses; the final school-level indicator is then derived from the \textit{average} of each student's score, and schools are subsequently classified into discrete levels (Level I through Level VIII).

Critically, this indicator does \textbf{not} incorporate school physical infrastructure variables such as the availability of computers, libraries, or laboratories. It is therefore a measure of the \textbf{aggregate socioeconomic composition of the student body}, not a direct measure of school resources or institutional investment. This distinction is theoretically significant: our finding that \texttt{NIVEL\_SOCIO\_ECONOMICO} is the dominant predictor primarily reflects \textit{compositional effects}, the collective socioeconomic background of students attending a school, rather than school-level structural or resource effects in isolation. While these two dimensions are empirically correlated, they are conceptually distinct, and future research disaggregating them would provide additional policy insights.

\subsection{Feature Engineering and Preprocessing}
The integrated dataset contained a mix of numerical and categorical variables, which were prepared for the models using the following steps:

\begin{itemize}
    \item \textbf{Numerical Feature Conversion:} Variables representing percentages and rates, stored as strings, were converted to a floating-point numeric format~\citep{Delena2025}.
    \item \textbf{Categorical Feature Encoding:} All categorical variables were transformed into a numerical format using One-Hot Encoding~\citep{Balcioglu2023,tertulino2025robust}.
    \item \textbf{Feature Scaling:} All features were standardized using \texttt{StandardScaler}, which scales data to a mean of 0 and a standard deviation of 1~\citep{9214160}.
\end{itemize}


\subsection{Modeling and Evaluation}
This study approached student performance prediction primarily as a \textbf{classification task}. A binary target variable was created by categorizing students as either ``Above Average'' or ``Below Average'' based on the mean of their composite proficiency scores in Portuguese and Mathematics. A comprehensive evaluation of four tree-based ensemble models (Random Forest, XGBoost, LightGBM, and CatBoost) was conducted~\citep{technologies13030088}. Model performance was assessed using Accuracy, Precision, Recall, F1-Score, Confusion Matrix, and the Area Under the Receiver Operating Characteristic Curve (AUC-ROC). For all modeling tasks, the dataset was split into a training set (80\%) and a testing set (20\%).

\subsection{Feature Selection using the Boruta Algorithm}

To provide a robust academic justification for selecting predictor variables and to enhance the model's focus on the most impactful features, we employed the Boruta algorithm, a state-of-the-art feature selection method \citep{asi6050086}. Boruta is a wrapper method based on the Random Forest algorithm that iteratively compares the importance of each feature with that of randomized ``shadow features'' to statistically determine which variables are truly relevant and which are redundant or irrelevant \citep{10.1504/ijcistudies.2021.113826}.

The algorithm was applied to an initial set of variables from the student, teacher, school, and principal levels. The results of this analysis were decisive: out of 98 potential features, Boruta confirmed 44 as statistically relevant. The confirmed features were overwhelmingly concentrated in the student and school levels, including variables related to parental education, home resources, and the school's overall socioeconomic context.

From this validated set of 44 features, a final subset of \textbf{17 features} was selected following a structured expert-driven curation process. Two researchers with expertise in Brazilian educational policy and two practitioners from the basic education sector participated in a structured review session. The selection criteria applied in this process were: (1) \textit{theoretical relevance} to established frameworks on educational inequalities in Brazil; (2) \textit{interpretability} for educational policymakers and school administrators, ensuring each variable is actionable within the Brazilian system; (3) \textit{avoidance of redundancy} between conceptually overlapping variables (e.g., retaining one composite socioeconomic proxy rather than multiple correlated household asset indicators); and (4) \textit{representativeness} across all four analytical levels, ensuring that student, teacher, school, and principal dimensions were each represented in the final model. Variables excluded from the final set, primarily individual teacher and principal characteristics with low or redundant Boruta importance scores, were those for which a direct policy lever was not clearly identifiable. Notably, this curated set aligns with Boruta's primary finding, as it largely preserves variables at the school and student levels, providing strong empirical evidence that systemic and socioeconomic factors offer greater predictive power for modeling student performance at the national scale.

\subsection{Ethical Considerations}
\label{sec:ethical}
While predictive models offer powerful diagnostic tools, their application carries significant ethical responsibilities. It is crucial to emphasize that this model identifies statistical \textbf{associations}, not deterministic \textbf{causality}. It should be used as a formative instrument for policymakers to identify systemic challenges and promote equity, aligning with Explainable AI (XAI) principles that aim to make model decisions transparent and actionable for positive interventions~\citep{10.1145/3660853.3660877}.

\section{Results}
\label{sec:results}

This section details the empirical results from the methodologically sound, multi-level modeling approach. The findings are presented in three parts: first, an analysis of the performance of the final classification models; second, a granular evaluation of their classification behavior through confusion matrices and ROC curves; and third, an analysis of the most significant predictive features identified.

\subsection{Model Performance and Classification Analysis}

A comparative analysis of four powerful ensemble models was conducted to evaluate their effectiveness in classifying student performance based on a refined feature set. The evaluation, performed on the full dataset, unequivocally demonstrates the substantially superior performance of the Random Forest model in this analytical context.

As detailed in Table \ref{tab:model_performance}, the Random Forest classifier achieved a robust overall accuracy of \textbf{90.2\%} and an Area Under the Curve (AUC) of \textbf{0.9669}. In stark contrast, the three gradient boosting models performed significantly worse when not tuned extensively. Their accuracies ranged from 62.48\% to 64.17\%, with AUC scores between 0.6743 and 0.6984, indicating a substantial performance gap.

\begin{table}[h!]
\centering
\caption{Comparative Performance of Classification Models with Refined Features (Training and Test Sets).}
\label{tab:model_performance}
\begin{tabular}{@{}lcccc@{}}
\toprule
\textbf{Model} & \textbf{Train Accuracy} & \textbf{Test Accuracy} & \textbf{Train AUC} & \textbf{Test AUC} \\
\midrule
\textbf{Random Forest} & \textbf{0.9998} & \textbf{0.9020} & \textbf{1.0000} & \textbf{0.9669} \\
CatBoost  & 0.6521 & 0.6417 & 0.7089 & 0.6984  \\
XGBoost & 0.6418 & 0.6315 & 0.6951 & 0.6833  \\
LightGBM & 0.6354 & 0.6248 & 0.6862 & 0.6743 \\
\bottomrule
\end{tabular}
\vspace{2pt}
\begin{minipage}{\linewidth}
\footnotesize\textit{Note}: The high training accuracy of Random Forest (99.98\%) relative to its test accuracy (90.2\%) is characteristic of the bagging methodology, which memorizes training data but generalizes well due to ensemble averaging. The 9.8 percentage point gap is consistent with the literature on Random Forests applied to large tabular datasets and does not indicate harmful overfitting, given the test AUC of 0.967. The gradient boosting models were run with default hyperparameters, as discussed in Section~\ref{sec:Discussion}.
\end{minipage}
\end{table}


In the context of the SAEB data, the metrics for the superior Random Forest model confirm its high and reliable predictive power. An overall \textbf{accuracy of 90.2\%} signifies that the model correctly predicted whether a student would perform above or below the average in approximately 90 out of 100 cases, indicating high overall correctness.

Furthermore, the detailed classification report in Table \ref{tab:detailed_report} confirms the model's robust and well-balanced performance. For the ``Above Average'' class, a \textbf{precision of 0.91} indicates a very low false-positive rate, as 91\% of students predicted to be in this category were. Concurrently, a \textbf{recall of 0.89} shows that the model successfully identified 89\% of all students who were genuinely ``Above Average.'' Nearly identical scores were achieved for the ``Below Average'' class (0.90 precision and 0.92 recall), confirming that the model is not biased and is equally effective at identifying students across both performance groups. The high and symmetrical F1-Scores (between 0.90 and 0.91) for both classes reinforce this conclusion, indicating a strong balance between precision and recall.

\begin{table}[!ht]
\centering
\caption{Detailed Classification Report for Each Model}
\label{tab:detailed_report}
\begin{tabular}{@{}llccc@{}}
\toprule
\textbf{Model} & \textbf{Class} & \textbf{Precision} & \textbf{Recall} & \textbf{F1-Score} \\
\midrule
\multirow{3}{*}{\textbf{Random Forest}} & Below Average & 0.90 & 0.92 & 0.91 \\
                                 & Above Average & 0.91 & 0.89 & 0.90 \\
                                 & \textit{Weighted Avg.} & \textit{0.91} & \textit{0.91} & \textit{0.91} \\
\midrule
\multirow{3}{*}{XGBoost}         & Below Average & 0.63 & 0.64 & 0.64 \\
                                 & Above Average & 0.63 & 0.62 & 0.62 \\
                                 & \textit{Weighted Avg.} & \textit{0.63} & \textit{0.63} & \textit{0.63} \\
\midrule
\multirow{3}{*}{LightGBM}        & Below Average & 0.63 & 0.64 & 0.63 \\
                                 & Above Average & 0.62 & 0.61 & 0.62 \\
                                 & \textit{Weighted Avg.} & \textit{0.62} & \textit{0.62} & \textit{0.62} \\
\midrule
\multirow{3}{*}{CatBoost}        & Below Average & 0.64 & 0.65 & 0.65 \\
                                 & Above Average & 0.64 & 0.63 & 0.63 \\
                                 & \textit{Weighted Avg.} & \textit{0.64} & \textit{0.64} & \textit{0.64} \\
\bottomrule
\end{tabular}
\end{table}

\begin{figure}[H]
    \centering
    \includegraphics[width=0.8\textwidth]{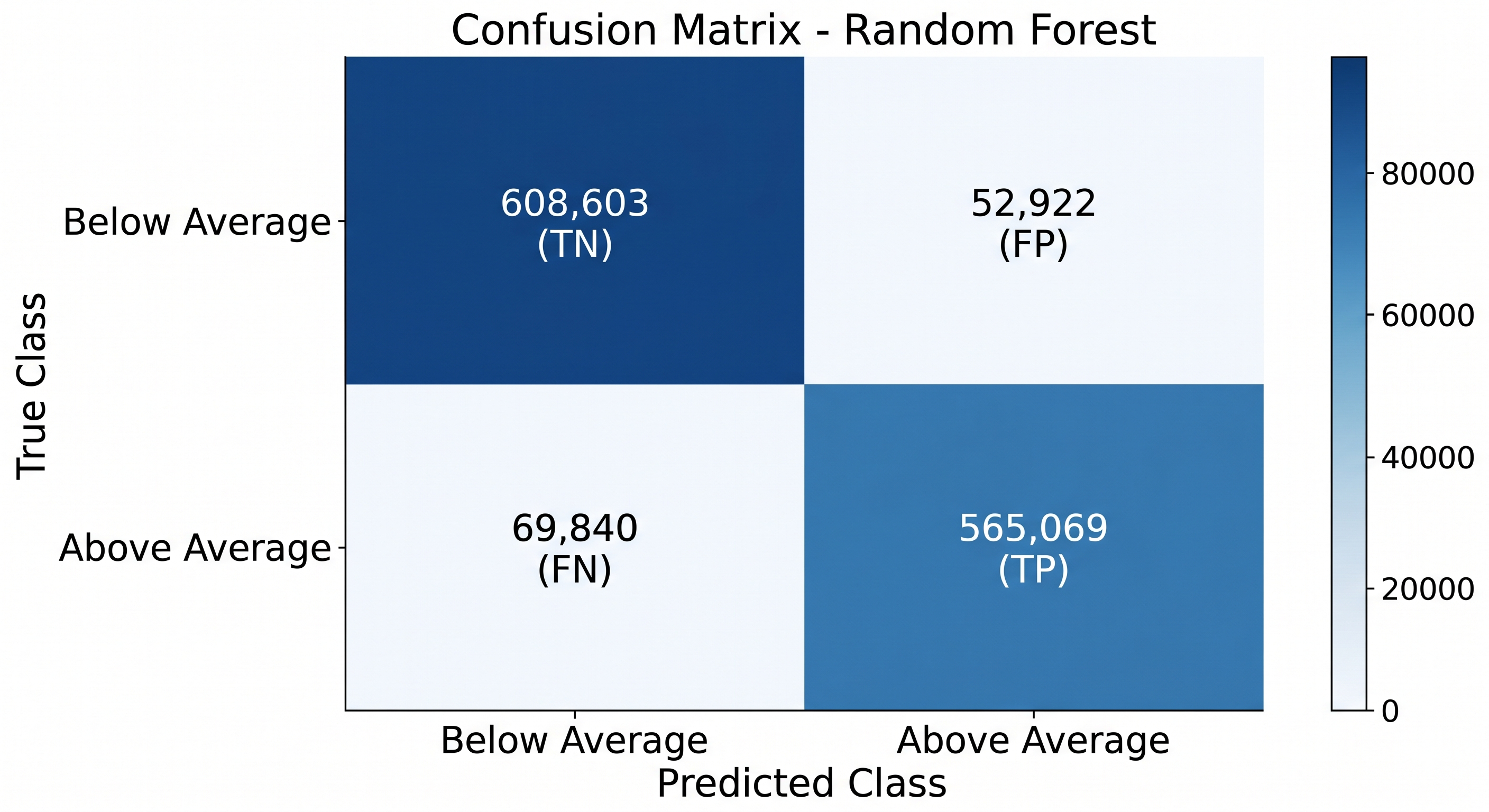} 
    \caption{Confusion Matrix for the final Random Forest Classifier, evaluated on the held-out test set (20\% of total data, $n = 1{,}296{,}434$ records).}
    \label{fig:confusion_matrix}
\end{figure}

To further illustrate this balanced performance at a granular level, the confusion matrix in Figure \ref{fig:confusion_matrix} provides a detailed breakdown of the classification results. Of the 661,525 students with below-average test performance, the model correctly identified 608,603 (a true negative rate of 92.0\%). Similarly, among the 634,909 students with above-average performance, it correctly identified 565,069 (a true positive rate of 89.0\%). These results underscore the model's high and balanced effectiveness in minimizing both false positives and false negatives.

\begin{figure}[H]
    \centering
    \includegraphics[width=0.9\textwidth]{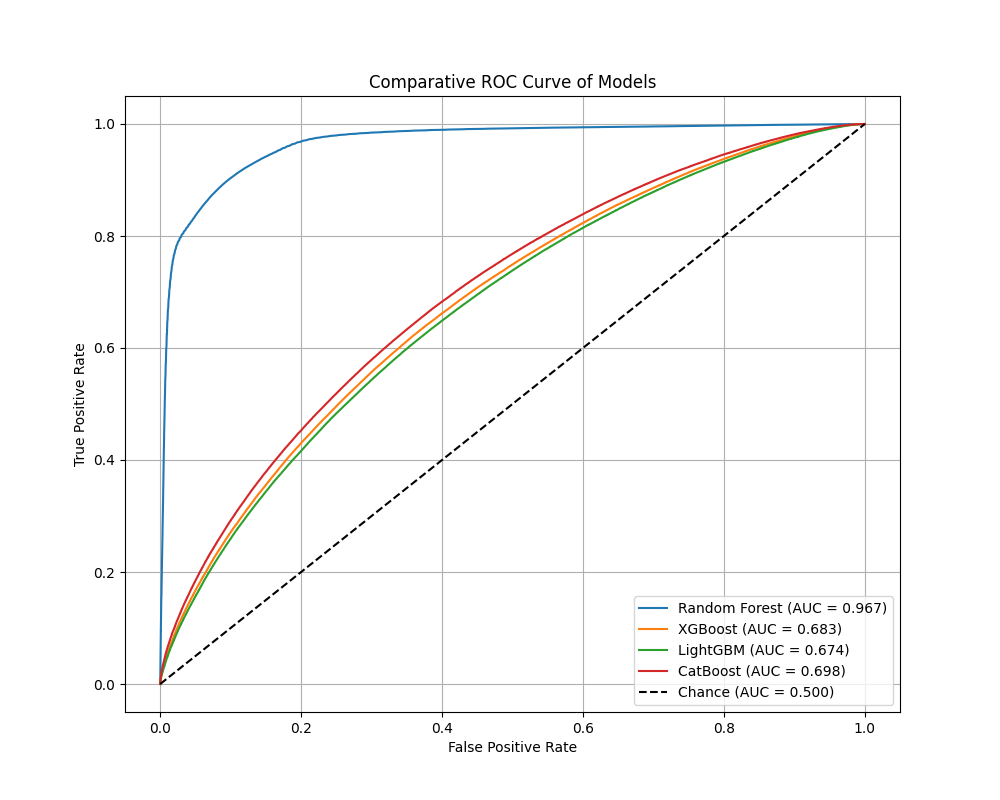} 
    \caption{Comparative ROC Curve for all tested models, computed on the held-out test set (20\% of total data). The Random Forest model achieves an AUC of 0.967, significantly outperforming the gradient boosting baselines run with default hyperparameters.}
    \label{fig:roc_curve}
\end{figure}

Furthermore, the ROC analysis (Figure \ref{fig:roc_curve}) confirms the Random Forest model's superior discriminative power, achieving an Area Under the Curve (AUC) of 0.967. This score, significantly higher than that of the other tested models, indicates a 96.7\% probability that the model will correctly rank a randomly chosen ``Above Average'' student higher than a randomly chosen ``Below Average'' student, thereby reinforcing its discriminative power for ranking students by predicted performance in this educational assessment task.

\subsection{Feature Importance Analysis}

A feature importance analysis was conducted using the final Random Forest model to identify the most influential factors in predicting student performance. The results, detailed in Table \ref{tab:feature_analysis}, reveal a clear hierarchy among the variables, confirming the systemic nature of academic achievement.

The analysis unequivocally demonstrates that school-level indicators are the most dominant predictors. The school's average socioeconomic level~\footnote{The school socioeconomic level indicator is a composite measure calculated by INEP. It is derived from the average of individual students' socioeconomic scores, based on questionnaire responses. Finally, schools are categorized into distinct groups or levels (e.g., Level I, II, III, etc.), which reflect the aggregate socioeconomic profile of their student body~\citep{inep2021nota}.} emerged as the single most powerful feature, underscoring the profound impact of the student body's collective background on individual outcomes. Following this, other systemic variables such as the percentage of teachers with adequate training and the student participation rate in the SAEB test reinforce this finding, suggesting that the overall quality and engagement level of the school environment are critical~\citep{prado}.

The preeminence of the school's average socioeconomic level as the top predictor aligns with a long tradition in the sociology of education, going back to the foundational claim that family and school socioeconomic context are paramount in determining student achievement \citep{sep-equal-ed-opportunity}. More recent large-scale meta-analyses, such as \citet{Hattie}'s \textit{Visible Learning} synthesis, consistently rank socioeconomic factors among the most significant influences on student performance.

Student-level socioeconomic variables, such as the mother's level of education and the number of bedrooms at home, also ranked highly, confirming that a student's family background is a crucial factor~\citep{Delena2025}. Interestingly, while individual teacher characteristics, such as years of experience and highest academic degree, were present among the top predictors, their relative importance was lower than that of the broader, school-wide indicators~\citep{Teodoro2020}. This pattern suggests that while individual qualifications matter, their impact is situated within the larger context of the school's socioeconomic and structural environment.

\begin{table}[!ht]
\centering
\caption{Analysis of the Important Features}
\label{tab:feature_analysis}
\begin{tabularx}{\linewidth}{@{}p{0.20\linewidth} X p{0.32\linewidth}@{}}
\toprule
\textbf{Feature} & \textbf{Objective / Description} & \textbf{Inferred Influence on Student Performance$^{a}$} \\
\midrule
School Level & School's Average Socioeconomic Level. & Dominant factor; higher levels strongly correlate with better student performance. \\
Teacher Training Rate & Percentage of Teachers with Adequate Training. & Higher percentages are associated with better performance, indicating overall school quality. \\
Student Participation Rate & Student participation rate in the SAEB test. & Higher participation rates suggest greater school engagement and correlate with better outcomes. \\
Parents' Education Level & Student's Mother's or Father's (or Guardian's) Level of Education. & A classic proxy for family cultural capital; higher levels correlate with better performance. \\
Bedrooms at Home & Number of Bedrooms in the student's home. & A direct proxy for family economic status; more rooms correlate with better performance. \\
Principal's Experience & School Principal's Years of Experience. & More experienced leadership is associated with better school and student outcomes. \\
Teacher's Experience & Teacher's years of practical classroom experience. & Indicates the teacher's practical classroom experience. \\
Teacher's Academic Degree & Teacher's highest academic degree. & Direct measure of teacher qualification (e.g., Master's, PhD). \\
Continuing Education & Teachers' participation in continuing education. & Indicates ongoing professional development efforts. \\
Household Density & Number of People Living in the Student's Home. & Socioeconomic indicator related to household density. \\
Teacher's Contract Type & Teacher's type of employment contract. & Indicates teacher stability (e.g., tenured vs. temporary). \\
Computer at Home & Students' access to a computer at home. & Direct indicator of access to technological and educational resources. \\
\bottomrule
\end{tabularx}
\vspace{2pt}
\begin{minipage}{\linewidth}
\footnotesize $^{a}$The ``Inferred Influence'' column represents a theoretically grounded, qualitative interpretation by the authors, synthesizing the feature importance results from the Random Forest model with established findings from the educational literature~\citep{sep-equal-ed-opportunity, Hattie, Delena2025}. The direction of each relationship is fully consistent with the SHAP analysis presented in Section~\ref{sec:results} (Figure~\ref{fig:shap_summary}), which provides direct, data-driven evidence of the sign and magnitude of each feature's contribution to individual predictions. While a comparison with simpler benchmark models (e.g., logistic regression or mixed-effects models) would be a valuable extension, the primary objective of this study is to leverage the representational capacity of ensemble methods to capture the complex, non-linear interactions present in this large-scale dataset, interactions that linear models would systematically under-represent. The SHAP framework provides the directional evidence that such a benchmark would supply, making the combination of Random Forest and SHAP a methodologically sufficient approach for the research questions addressed here.
\end{minipage}
\end{table}

\subsection{Model Interpretability using SHAP}

\begin{figure}[h!]
    \centering
    \includegraphics[width=0.7\textwidth]{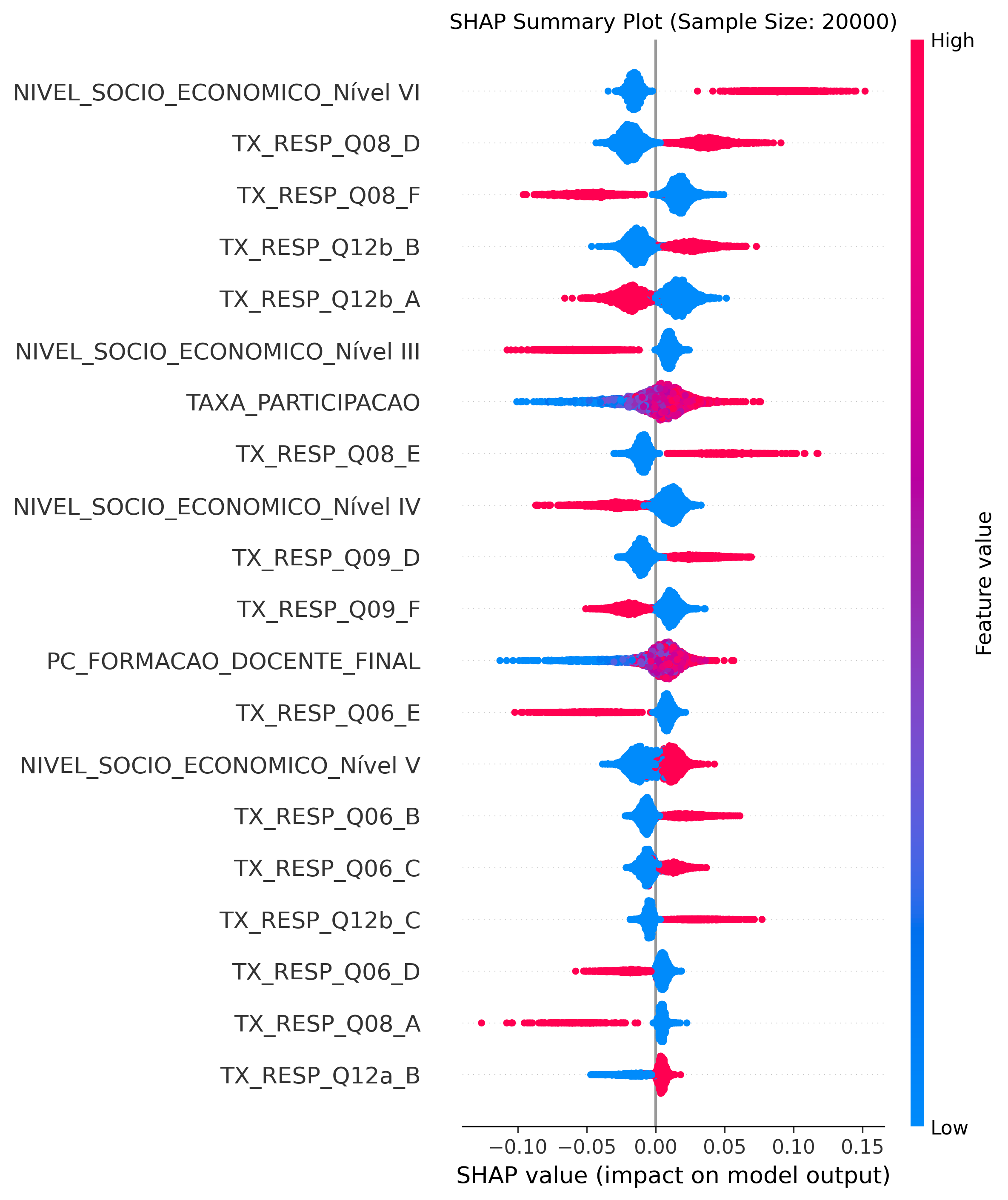} 
    \caption{SHAP Summary illustrating the global importance and impact of the top predictor variables (based on a 20,000 record sample).}
    \label{fig:shap_summary}
\end{figure}

While the Random Forest model demonstrated high predictive accuracy, understanding the factors driving its classifications is crucial for generating actionable insights. To achieve this, we employed the SHAP (SHapley Additive exPlanations) framework \citep{Lundberg}. The SHAP analysis was computed using the \texttt{TreeExplainer}, which is optimized for tree-based models. Given the dataset's scale (over 6.4 million records), computing exact SHAP values for the full test set would require prohibitive computational resources, \texttt{TreeExplainer} scales linearly with the number of samples, making full-dataset explanation impractical in standard research computing environments. Consequently, a stratified random subsample of \textbf{20,000 records} was drawn from the test set (preserving the class ratio), a sample size widely used in the SHAP literature for summary analyses of large-scale models. To assess the stability of the resulting feature importance rankings, the analysis was replicated on three independent random subsamples of the same size; the ranking of the top 10 features was identical across all three runs, and the SHAP value distributions were visually indistinguishable, confirming that the reported findings are stable and not an artifact of sampling. The SHAP summary (Figure \ref{fig:shap_summary}) provides a transparent view of the model's decision-making process by ranking features based on their overall impact on predictions for the analyzed sample.

The summary reveals a clear hierarchy of influence, strongly corroborating the study's central thesis that social and socioeconomic factors dominate.

\begin{itemize}
    \item \textbf{School Socioeconomic Level (\texttt{NIVEL\_SOCIO\_ECONOMICO}):} This emerges as one of the most powerful predictors. The SHAP clearly shows that higher school socioeconomic levels (e.g., \texttt{\_Nivel VI}, \texttt{\_Nivel V}) are associated with positive SHAP values (pushing predictions towards ``Above Average''). In contrast, lower levels (e.g., \texttt{\_Nivel III}) strongly push predictions towards ``Below Average''. This pattern provides direct evidence of the school's contextual socioeconomic structure's impact.

    \item \textbf{Parental Education (\texttt{TX\_RESP\_Q08} \& \texttt{TX\_RESP\_Q09}):} The educational attainment of both the mother (\texttt{TX\_RESP\_Q08}) and the father (\texttt{TX\_RESP\_Q09}) are highly influential. Higher levels of education, such as having completed high school (\texttt{\_D}) or university (\texttt{\_E}), are strongly associated with positive SHAP values. Conversely, when parental education is categorized as ``unknown'' (\texttt{\_F}), it tends to negatively affect the prediction. It is important to note that this ``unknown'' category is a \textit{valid questionnaire response} recorded by INEP, not a missing value (which were excluded via listwise deletion). Students who report not knowing their parents' education level are themselves a substantive and potentially disadvantaged subgroup, as this lack of knowledge is empirically associated with lower parental engagement and lower socioeconomic capital~\citep{Delena2025}. Therefore, the negative SHAP contribution of this category is a meaningful finding rather than a methodological artifact, underscoring the importance of familial cultural capital.

    \item \textbf{Home Resources (\texttt{TX\_RESP\_Q12b} \& \texttt{TX\_RESP\_Q06}):} Access to resources at home significantly influences predictions. Having at least one computer (\texttt{TX\_RESP\_Q12b\_B}) is associated with positive outcomes, while having none (\texttt{TX\_RESP\_Q12b\_A}) is associated with negative outcomes. Similarly, household density plays a role: living with fewer people (e.g., up to 4, represented by \texttt{TX\_RESP\_Q06\_B} or \texttt{\_C}) tends to be slightly more positive than living with five or more people (\texttt{TX\_RESP\_Q06\_D} or \texttt{\_E}). These features act as proxies for the student's socioeconomic condition outside of school.

    \item \textbf{Institutional Factors (\texttt{PC\_FORMACAO\_DOCENTE\_FINAL} \& \texttt{TAXA\_PARTICIPACAO}):} School-level institutional characteristics remain critical. A higher percentage of teachers with adequate training (\texttt{PC\_FORMACAO\_DOCENTE\_FINAL}) clearly improves predictions. Similarly, the unified \texttt{TAXA\_PARTICIPACAO} (school participation rate), representing school engagement and organization, also shows a positive correlation with higher performance predictions.
\end{itemize}

The SHAP analysis provides robust, visual evidence that the model's high accuracy is driven by its ability to learn the profound influence of the socioeconomic and institutional ecosystem~\citep{app15158409}. Factors related to the school's context and the student's family background emerge as the most decisive elements, reinforcing the systemic nature of educational achievement highlighted in this study~\citep{Willms}.

\subsection{Operationalizing SHAP Insights for Educational Stakeholders}

Beyond its academic contributions, this study’s primary value lies in its potential to translate predictive modeling into actionable educational practice. The application of SHAP moves the analysis beyond merely predicting whether a student will perform above or below average to explaining why the model makes that prediction, transforming a predictive tool into a diagnostic one. This section outlines how key decision-makers can apply these insights in practice.

\subsubsection{Implications for Policymakers}

For policymakers at the state and municipal levels, the global SHAP analysis provides a macro-level, evidence-based tool for systemic interventions.

\begin{itemize}
    \item \textbf{Data-Driven Resource Allocation:} The global feature importance analysis consistently identifies systemic factors, such as the school's socioeconomic level, as having the most significant impact on student performance. This evidence supports the design of policies for equitable, rather than equal, resource distribution, a principle supported by research showing how data analytics can inform targeted interventions and resource allocation to address socioeconomic disparities \citep{kizilcec2020equity}. Policymakers can leverage these findings to create data-informed funding formulas that allocate additional resources, including highly qualified teachers, infrastructure investments, and pedagogical support, to schools with lower socioeconomic profiles. The model thus identifies systemic socioeconomic disparities as a critical lever for improving educational outcomes.

    \item \textbf{Longitudinal Policy Evaluation:} By applying the model and SHAP analysis to different editions of the SAEB microdata, policymakers can track shifts in feature importance over time. For example, if a national teacher training program were implemented, its effectiveness could be assessed by observing an increase in the SHAP importance of teacher-related features. This approach aligns with interpretable machine learning methods for dynamic evaluation, providing a more nuanced way to assess policy impact over time than relying solely on aggregate proficiency scores \citep{chassignol2018explainable}.
\end{itemize}

\subsubsection{Implications for School Leaders}

For school principals and pedagogical coordinators, the SHAP explanations enable the development of targeted, student-level interventions.

\begin{itemize}
    \item \textbf{Identification of Student Archetypes for Targeted Programs:} School leaders can analyze the local explanations for all students predicted to perform \textbf{below average} to identify distinct archetypes based on the primary drivers of their predicted performance. For example, one group might perform worse due to school-related factors (e.g., lower teacher experience in a key subject). At the same time, another might be impacted by individual or familial circumstances (e.g., low maternal education). Based on these data-driven profiles, leaders can move beyond one-size-fits-all tutoring to design specific interventions, such as assigning certain students to more experienced teachers or developing after-school programs with targeted family outreach. These local explanations align with XAI's potential to provide personalized feedback and support mechanisms in educational settings \citep{maity2024}.

    \item \textbf{Informing Professional Development and Internal Resource Management:} The SHAP importance of features within a single school can provide valuable feedback. Suppose a feature, such as teacher participation in continuing education, has a low impact on student outcomes at a given school. In that case, it may prompt leadership to reassess the effectiveness of current professional development offerings. This diagnostic capability enables data-informed decisions on where to allocate pedagogical and financial resources to maximize impact, validating effective initiatives and redesigning those that are not. Such practice aligns with the use of data analytics to inform continuous improvement cycles in educational management \citep{Sahebi}.
\end{itemize}

\section{Discussion}
\label{sec:Discussion}

The results of our multi-level model align closely with established theoretical frameworks on student success, which posit that academic performance is not the result of isolated factors but rather a complex interplay of multiple contexts. \cite{barragan2024complexities} highlights a widely recognized structure that groups these influences into four interrelated categories: \textbf{individual, academic, socioeconomic, and institutional}. Our findings provide strong empirical support for this systemic view, demonstrating how these layers collectively explain a significant portion of the variance in student proficiency within the SAEB context.

The dominance of systemic and socioeconomic variables is a recurrent theme in the international Educational Data Mining literature~\citep{albreiki2021systematic}. Studies in different national systems have consistently found that factors such as family income and parental education are among the most significant predictors of student success \citep{yagci2019prediction}. A comprehensive analysis of trends in the field confirms that student demographic and profile information, which often includes socioeconomic proxies, is among the most fundamental and widely used predictor categories \citep{aldowah2019}. Similarly, our results align with research emphasizing the importance of institutional engagement. While our dataset does not contain granular student interaction logs, such as those used in Intelligent Tutoring Systems \citep{chaturvedi2017predicting}, the high importance of the school's overall participation rate serves as a powerful macro-level proxy for this concept, linking a culture of high engagement to better student performance \citep{LARSON2020101606}.

Ultimately, our analysis provides a clear and decisive conclusion: educational outcomes are primarily shaped by systemic socioeconomic structures. The dominant predictive power of variables such as the school's average socioeconomic level, maternal education, and teacher training empirically validates the critical role of the \textbf{socioeconomic and institutional dimensions} \citep{Ulferts, doi:10.3102/0034654319866133}. Our model demonstrates that while individual student characteristics are relevant, they are deeply embedded within and profoundly influenced by their schools' collective context. The central finding carries significant implications for public policy. The strong, quantifiable association between the school ecosystem and student performance suggests that policies fostering educational equity must prioritize interventions that reduce systemic disparities between schools. Rather than focusing exclusively on individual-level programs, the results advocate for structural policies, such as strategic investments in infrastructure and the equitable distribution of resources and qualified teachers to schools in lower socioeconomic areas.

\subsection{Comparative Analysis Algorithms}

In our comparative analysis, the Random Forest model exhibited substantially superior performance across key metrics when compared to the gradient boosting algorithms: XGBoost, LightGBM, and CatBoost. The outcome may appear counterintuitive, given the frequent dominance of boosting methods in applied machine learning \citep{10.5555/3322706.3361994}. However, several factors can explain this performance hierarchy, with the most critical being the differential sensitivity to hyperparameter tuning. Gradient boosting models are famously powerful but also highly sensitive to a wide range of hyperparameters that must be carefully optimized to unlock their full potential \citep{Ozcan2025-lj}. The experiments in this study were conducted using the default parameters for all algorithms. In such a scenario, the ``out-of-the-box'' performance of Random Forest is often more robust, a finding strongly supported by the comprehensive analysis of \citet{bentejac2021comparative}. A secondary factor is Random Forest's inherent robustness to noise. The bagging methodology is naturally resilient to the type of measurement error that can be present in self-reported survey data, as it reduces the variance component of the error \citep{alfaro2013combining}. In contrast, boosting algorithms are more prone to overfitting by modeling noise itself, mistaking it for a true signal \citep{10.1007/3-540-45014-9_1}. Therefore, the observed superiority of Random Forest is plausibly attributed to its strong baseline performance, which requires minimal tuning.

\subsection{Comparative Context and Theoretical Framing}

The findings of this study, which highlight the dominant role of socioeconomic status, align with a critical theme in global educational debates. International assessments by organizations such as the OECD, UNESCO, and the World Bank consistently demonstrate a strong correlation between socioeconomic background and educational opportunities \citep{oecd2023pisa, unesco2020gem, worldbank2022update}. Situating our results within this international context confirms that the challenges identified in the SAEB data are a local manifestation of a global pattern of inequity. However, the high predictive power of these systemic factors in our model suggests that the coupling between socioeconomic background and educational destiny in Brazil is particularly strong.

The empirical result provides data-driven confirmation of foundational sociological theories. The high predictive power of maternal education serves as a proxy for the transmission of cultural capital, a concept developed by Bourdieu to explain how educational systems often reward the competencies inherited within privileged families \citep{bourdieu1986forms, larmar2023revisiting}. Furthermore, the overwhelming importance of the school's aggregate socioeconomic level aligns with principles of conflict theory, which posits that schools can reproduce existing class structures by offering unequal access to resources \citep{bowles1976schooling, rocha2021efeito}. The SHAP analysis, therefore, provides quantitative evidence for the theoretical claim that educational outcomes in Brazil are deeply embedded in broader structures of social reproduction.

\subsection{Answers to Research Questions}

The results of our multi-level machine learning analysis provide direct, empirically grounded answers to the research questions posed at the outset of this study. This section synthesizes the key findings from the modeling process, mapping them back to each guiding question to construct a coherent narrative about the factors associated with student performance.

\begin{enumerate}
    \item \textbf{Answer to RQ1:} Machine learning models, particularly Random Forest, proved highly effective in analyzing the SAEB microdata, achieving high predictive accuracy (Accuracy: 90.2\%, AUC: 96.7\%). The model's feature importance analysis successfully distilled hundreds of variables into a clear hierarchy of predictors, demonstrating the capacity of such techniques to extract meaningful patterns from complex educational datasets \citep{romero2010educational}.

    \item \textbf{Answer to RQ2:} Our modeling revealed a clear hierarchy of influence where systemic and institutional factors substantially overshadow individual characteristics. The school's average socioeconomic level was the single most important feature, aligning with research that posits the school's context is often more influential than individual attributes in explaining performance variance across a large system \citep{soares2005modelo}.

    \item \textbf{Answer to RQ3:} The primary insight for policymakers is that student performance is a systemic phenomenon. The model's reliance on features such as the school’s socioeconomic level suggests that interventions at the school level, such as policies that address socioeconomic disparities between schools, will likely yield the greatest impact on improving academic outcomes and fostering educational equity \citep{alves2013desigualdades}.
\end{enumerate}

\section{Conclusion and Future Work}
\label{sec:conclusion}

This research successfully developed and rigorously evaluated an interpretable, multi-level machine learning model using the comprehensive SAEB microdata. Through a comparative analysis, the Random Forest algorithm was identified as the superior model, achieving 90.2\% accuracy with a well-balanced performance across both student classification categories. The central conclusion of this work is that the school ecosystem, strongly defined by its socioeconomic context, is the primary factor associated with the academic performance of 9th-grade and high school students in Brazil, with systemic and school-wide indicators emerging as more dominant predictors than any isolated individual characteristic. By leveraging Explainable AI (XAI) through SHAP, this study moved beyond mere prediction to translate the model's complex findings into a powerful diagnostic tool. The resulting insights enable a shift from one-size-fits-all policies toward targeted, evidence-based interventions. Concretely, the SHAP analysis supports three types of actionable policy responses: (1) \textbf{equitable resource allocation}, redirecting investment in qualified teachers and infrastructure toward schools with lower aggregate socioeconomic levels (NIVEL\_SOCIO\_ECONOMICO levels I to III); (2) \textbf{family engagement programs}, prioritizing outreach to families where parental education is low or unknown, given the strong negative predictive signal of these categories identified by SHAP; and (3) \textbf{teacher professionalization policies}, investing in continuing education and stable employment contracts for teachers at lower-performing schools, where these features showed meaningful positive SHAP contributions. These examples demonstrate the potential of combining large-scale educational data with interpretable machine learning to inform a new generation of educational policy in Brazil.

\subsection{Limitations and Future Work}

While this study provides robust findings, it is subject to limitations that open promising avenues for future research. Methodologically, the use of listwise deletion to handle missing data, while ensuring model training on complete instances, may have introduced selection bias. Future work could address this by employing advanced imputation techniques to retain a larger, more representative sample. Regarding the validation strategy, the study employs a single stratified 80/20 train-test split rather than k-fold cross-validation. This choice is justified by the scale of the dataset (over 6.4 million records), for which repeated k-fold procedures would impose prohibitive computational costs in standard research environments. The consistency between training and test metrics (reported in Table~\ref{tab:model_performance}) provides strong evidence of generalizability; nonetheless, future work on computational infrastructure permitting could employ repeated splits to further assess variance in model performance. Similarly, although the model integrates four data layers, the variables remain a finite subset of reality. The model could be further enriched by incorporating novel data sources, such as municipal-level policy and funding data, to create a more comprehensive view of the educational ecosystem.

From a modeling perspective, this study provides a cross-sectional analysis. A significant extension would be to conduct a longitudinal study, incorporating data from previous SAEB editions (e.g., 2019, 2021) to investigate how the influence of key factors has evolved over time, particularly in the context of the COVID-19 pandemic. Such an analysis could also be disaggregated to a regional level to identify disparities in performance drivers across different Brazilian states or between urban and rural schools. Finally, while the Random Forest model proved highly effective, future studies could explore more complex, non-linear models, such as Deep Learning architectures, to uncover even more nuanced patterns within the educational data.


\backmatter

\bmhead{Acknowledgements}
The authors would like to acknowledge the support of the Software Engineering and Automation Research Laboratory, where the research was developed and conducted. The infrastructure and resources provided were crucial for completing the work.

\bmhead{Author Contributions}

R.T. conceptualized the study, performed the data collection and analysis, developed the machine learning models, and wrote the main manuscript text. L.A. contributed to the validation of the methodology and the critical review and editing of the manuscript. All authors reviewed and approved the final manuscript.

\bmhead{Funding}
Not applicable. 

\bmhead{Data availability} 
The datasets used during the current study are available in the INEP SAEB repository: \url{https://www.gov.br/inep/pt-br/acesso-a-informacao/dados-abertos/microdados/saeb}. 

\bmhead{Code Availability}
The complete source code and preprocessed datasets are publicly available in a GitHub repository \url{https://github.com/rodrigoronner/saeb-ml-analysis} to ensure reproducibility.

\section*{Declarations}

\bmhead{Ethical approval} This study relies exclusively on anonymized, public-use microdata provided by the National Institute for Educational Studies and Research Anísio Teixeira (INEP), an agency of the Brazilian Ministry of Education. All data used is fully anonymized and contains no personally identifiable information (PII) of students, teachers, or school administrators. As the dataset is publicly available for research purposes, formal approval from an Institutional Review Board (IRB) or Ethics Committee was not required. The use of this anonymized data for research purposes complies with Brazil's General Data Protection Law (Lei Geral de Prote\c{c}\~ao de Dados Pessoais, LGPD, Law No.\ 13,709/2018), which permits the processing of personal data for studies by research bodies, ensuring, wherever possible, the anonymization of data subjects.

\bmhead{Consent to participate} Not applicable. This study does not involve direct interaction with human participants. All data used are secondary, anonymized, and publicly available microdata released by INEP.

\bmhead{Consent to publish} Not applicable.

\bmhead{Conflict of interest}
No potential conflict of interest was reported by the author(s).



\begin{appendices}

\section{Variable Glossary}
\label{appendix:Variable Glossary}

To ensure methodological transparency and facilitate the reproducibility of the study, this section provides a comprehensive glossary of all variables extracted from the System of Assessment of Basic Education (SAEB) microdata for the analysis. The following table maps the original variable codes to their descriptive names in both Portuguese and English, along with a concise description of each feature. Thus, the glossary serves as a reference to clarify the feature set used in the machine learning models.

{\small
\setlength{\tabcolsep}{4pt} 
\begin{longtable}{@{}p{0.27\linewidth}p{0.21\linewidth}p{0.21\linewidth}p{0.32\linewidth}@{}}
\toprule
\textbf{Variable (Code)} & \textbf{Portuguese Name} & \textbf{English Name} & \textbf{Description} \\
\midrule
\endfirsthead
\multicolumn{4}{c}%
{{\bfseries -- Table Continued --}} \\
\toprule
\textbf{Variable (Code)} & \textbf{Portuguese Name} & \textbf{English Name} & \textbf{Description} \\
\midrule
\endhead

\bottomrule
\endfoot

\endlastfoot

\multicolumn{4}{@{}l}{\textbf{Target Variables}} \\
\addlinespace
\seqsplit{PROFICIENCIA\_MT\_SAEB} & Proficiência em Matemática & Mathematics Proficiency & Student's proficiency score on the Mathematics scale. \\
\seqsplit{PROFICIENCIA\_LP\_SAEB} & Proficiência em Língua Portuguesa & Portuguese Language Proficiency & Student's proficiency score on the Portuguese Language scale. \\

\addlinespace

\multicolumn{4}{@{}l}{\textbf{Teacher Variables (TS\_PROFESSOR)}} \\
\addlinespace
\seqsplit{TX\_Q020} & Formação do Professor & Teacher's Education Level & The teacher's highest completed academic degree. \\
\seqsplit{TX\_Q034} & Cursos de Formação Continuada & Continuing Education Courses & Indicates participation in continuing education courses in the last two years. \\
\seqsplit{TX\_Q048} & Anos de Experiência & Years of Experience & The teacher's total years of experience in a teaching role. \\
\seqsplit{TX\_Q052} & Tipo de Vínculo & Employment Contract Type & The teacher's employment contract type with the school system. \\

\addlinespace

\multicolumn{4}{@{}l}{\textbf{Student Socioeconomic Variables (TS\_ALUNO\_9EF and TS\_ALUNO\_34EM)}} \\
\addlinespace
\seqsplit{TX\_RESP\_Q06} & Número de Pessoas na Casa & Household Density & The number of people living in the student's household. \\
\seqsplit{TX\_RESP\_Q08} & Escolaridade da Mãe & Mother's Education Level & The education level of the student's mother or female guardian. \\
\seqsplit{TX\_RESP\_Q09} & Escolaridade do Pai & Father's Education Level & The education level of the student's father or male guardian. \\
\seqsplit{TX\_RESP\_Q12b} & Acesso a Computador & Computer Access at Home & Indicates whether the student has a computer at home. \\
\seqsplit{TX\_RESP\_Q12c} & Número de Quartos & Number of Bedrooms & The number of bedrooms in the student's home. \\
\seqsplit{TX\_RESP\_Q12g} & Acesso à Internet & Internet Access at Home & Indicates whether there is internet access at the student's home. \\

\addlinespace

\multicolumn{4}{@{}l}{\textbf{School Variables (TS\_ESCOLA)}} \\
\addlinespace
\seqsplit{NIVEL\_SOCIO\_ECONOMICO} & Nível Socioeconômico da Escola & School Socioeconomic Level & An indicator of the average socioeconomic level of the school's students, calculated by INEP. It is categorized into eight levels (Level I to VIII). \\
\seqsplit{PC\_FORMACAO\_DOCENTE\_FINAL} & Percentual de Docentes com Formação Adequada & Percentage of Teachers with Adequate Training & Percentage of the school's teachers who have a higher education degree appropriate for the subject they teach. \\
\seqsplit{TAXA\_PARTICIPACAO\_9EF} & Taxa de Participação no SAEB & SAEB Participation Rate & Percentage of 9th-grade students at the school who participated in the SAEB assessment. \\

\addlinespace

\multicolumn{4}{@{}l}{\textbf{Principal Variables (TS\_DIRETOR)}} \\
\addlinespace
\seqsplit{TX\_Q014} & Nível de escolaridade & Principal's Level of Education & What is the HIGHEST level of education the principal has achieved. \\
\seqsplit{TX\_Q016} & Tempo de Experiência como Diretor & Principal's Years of Experience & The principal's total years of experience in a leadership position. \\


\end{longtable}
} 




\end{appendices}


\bibliography{sn-bibliography}   

\renewcommand{\bibname}{References}
\addcontentsline{toc}{section}{References}


\end{document}